\documentclass[11pt]{article}

\usepackage[table]{xcolor}
\usepackage[final]{acl}

\usepackage{times}
\usepackage{latexsym}

\usepackage[T1]{fontenc}

\usepackage[utf8]{inputenc}

\usepackage{inconsolata}

\usepackage{graphicx}

\usepackage{booktabs}
\usepackage{adjustbox}
\usepackage{multirow}
\usepackage{array}
\usepackage{rotating}
\usepackage{amssymb}
\usepackage{tikz}
\usepackage{tikzpeople}
\usepackage{utfsym}
\usetikzlibrary{calc,shapes.callouts}
\usepackage{gb4e}
\noautomath

\title{Modal Sense Disambiguation in Language Models: The Role of Context, Training Domain, and Input Frequency}

\author{
Ezgi Ba\c{s}ar \quad
Yevgen Matusevych \quad
Annemarie van Dooren \quad
Arianna Bisazza \\
Center for Language and Cognition (CLCG), University of Groningen \\
\texttt{\{e.basar, yevgen.matusevych, a.m.f.van.dooren, a.bisazza\}@rug.nl}
}

\begin{document}

\maketitle

\begin{abstract}

Modals such as \textit{may} and \textit{must} can take on different senses depending on discourse context, alternating between root interpretations (e.g., permission, obligation) and epistemic interpretations (e.g., knowledge, belief). While modals are known to be notoriously difficult for children to acquire, limited research exists on whether language models (LMs) can reliably distinguish between different modal senses. We evaluate seven LMs that vary in architecture, size, and training data type using vector-based similarity and surprisal measures across four datasets. Specifically, we investigate whether LMs can leverage contextual information to disambiguate modal senses and whether their performance reflects the distribution of modal senses in their training data. We show that larger LMs trained on diverse text successfully leverage additional contextual information when available. In the absence of context, they tend to favor the dominant sense based on corpus statistics. Smaller developmentally plausible LMs, including those trained on child-directed language (CDL), fail to show consistent sensitivity to modal sense distinctions and do not benefit from context to the same degree.\footnote{Code is available at https://github.com/ezgibasar/modal-sense-disambiguation-in-lms.}

\end{abstract}

\section{Introduction}

\begin{figure}
    \centering
    \includegraphics[width=0.46\textwidth]{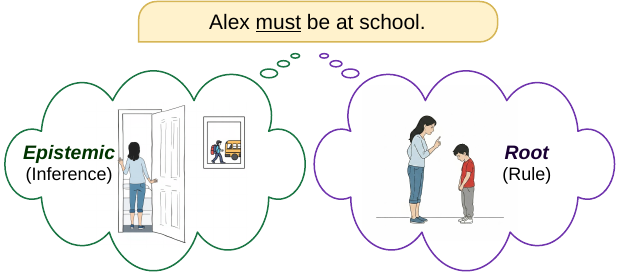}
    \caption{An example of modal sense disambiguation.}
    \label{fig:scenario}
\end{figure}

A sentence like ``Alex must be at school'' can be interpreted in two distinct ways (Figure \ref{fig:scenario}). In the first case, we can imagine a situation in which Alex's parents do not permit him to skip school, hence conveying a sense of obligation. In the other case, we can picture a parent entering Alex's room in the morning and finding it empty, inferring that Alex is likely at school. Here, the same sentence expresses a belief about where Alex is. This type of ambiguity is common with modals such as \textit{may} and \textit{must}, whose interpretation depends on the linguistic and non-linguistic context. 
Linguists distinguish two broad categories of modal senses, or ``flavors''. Root sense expresses notions that have to do with permission, obligation and ability, as in our first example. Epistemic sense expresses notions related to knowledge and belief, as exemplified in the second scenario.

Ambiguous modal expressions are ubiquitous in natural language. Roughly half of the world's languages feature a modal expressing root and epistemic senses \citep{vanderauwera2005modal}. Nevertheless, modals are notoriously complex for children to acquire due to their context-dependency and lack of physical grounding \citep{cournane2020leaving, vandooren2022figuring}.

In natural language processing (NLP), distinguishing between modal senses can be seen as a subtype of the more general task of word sense disambiguation \cite{marasovic2016multilingual}. Word sense disambiguation systems have long been developed to grapple with the polysemy observed in natural language. Modals add an extra layer of complexity for these systems, because it is possible for identical sentences to receive different interpretations based on the conversational context \citep{Kratzer1991}. This property makes modals highly challenging for NLP systems, including language models (LMs).

In this study, we ask whether and how various LMs distinguish between different modal senses in English. More specifically, we address three questions: \textbf{(1)} whether additional conversational context aids LMs in disambiguating modal senses, \textbf{(2)} whether LMs' discrimination of modal senses generalizes to out-of-domain evaluation data or is confined to their training domain, and \textbf{(3)} whether an LM's performance reflects the distribution of modal senses in its training input.

To answer these questions, we consider seven LMs of varying architectures, sizes and training data types. Alongside LMs pretrained on large amounts of diverse text data, we evaluate smaller LMs pretrained on child-directed language (CDL) for a more developmentally plausible reference point, including both masked and causal architectures.
We evaluate these LMs on our adaptations of three existing datasets \citep{zhou-etal-2015-semantically, tulling2020neural, vandooren2017learning}. We first probe vector representations of modals to investigate if LMs have distinct representations for modal senses.
We then use surprisal measures in a minimal-pair setup to assess whether LMs show a preference for one modal sense over another.

This study provides a comprehensive analysis of LMs' ability to discriminate between modal senses, accounting for both larger and smaller developmentally plausible LMs. Furthermore, we contribute adapted evaluation datasets in which the availability of contextual cues is manipulated, enabling a controlled investigation into the role of context in modal sense disambiguation. Finally, we explore whether input frequency affects LMs' behavior, offering a potential bridge between computational modeling and linguistic theories of modal acquisition.

\section{Background}

\subsection{Modal senses in NLP}

Prior work involving modal senses has largely focused on annotation and classification. \citet{ruppenhofer2012yes} developed an annotation scheme for English modals distinguishing root and epistemic uses. \citet{rubinstein-etal-2013-toward} created a more fine-grained framework capturing sense, strength, evidence, and attitude. \citet{zhou-etal-2015-semantically} explored paraphrase-driven sense projection and semantically enriched classification models, achieving strong performance across genres.
\citet{li2019modal} demonstrated that basic SVM classifiers improve when word embedding features are used for modal sense classification. \citet{dehouck2023revisiting} revisited the task using contextual word embeddings, employing BERT as a feature extractor for the token associated with the modal. \citet{marasovic2016multilingual} used a convolutional neural network architecture for modal sense classification and identified useful features such as past versus non-past tense, dynamic versus stative verbs, passive constructions, negation, and telic clauses. \citet{colli2024modal} presented the first modal sense classification work for French, fine-tuning a range of BERT variants on annotated sentences with augmented context. Notably, these studies are all classification tasks relying on supervised learning from labeled samples rather than examining the model-internal representations of pretrained LMs.

Most relevant to our work is \citet{wagner2023probing}, who investigated BERT's ability to encode sentence modality and modal sense across varieties of English. They trained probing classifiers on contextualized embeddings of modals and sentence embeddings, also testing BERT's ability to predict masked modals. However, probing classifiers can only reveal what information is extractable from a representation, and they do not provide direct evidence about whether the model uses that information \citep{hewitt2019designing}; therefore, we adopt an alternative methodology that combines vector similarity probing with surprisal-based measures.

It is worth noting that similar methods have been employed in previous word sense disambiguation research. \citet{nair-etal-2020-contextualized} demonstrate that distances between sense centroids in contextualized vectors significantly correlate with human judgments, providing support that vector-based similarity can capture human-like meaning distinctions. Moreover, \citet{temerko-etal-2025-continuous} report that surprisal scores correlate with human acceptability ratings for attested polysemes. Given our focus on developmental plausibility, these methods are also better aligned with our objectives than probing classifiers.

To our knowledge, no prior work has systematically compared causal and masked LMs for modal sense disambiguation. Causal LMs, whose auto-regressive objective more closely approximates the incremental nature of human language processing \citep{goldstein2022shared}, remain underexplored in this context. Our work addresses this gap by probing the vector representations and predictive behavior of a range of LMs, including both causal and masked architectures. We also examine CDL-trained LMs to investigate whether and how LMs learn different modal sense representations from  more developmentally plausible input.

\subsection{Acquisition of modals}

The acquisition of modals presents well-documented difficulties for children, especially for the epistemic sense. Several explanations for this challenge have been proposed in the developmental literature. One account appeals to Theory of Mind (ToM), suggesting that epistemic modals require the ability to attribute mental states to others, which develops around age four \citep{papafragou1998acquisition}. However, studies using non-verbal measures have found evidence of ToM-related abilities in infants as young as 15 months \citep{onishi2005do}, and two-year-olds appropriately use epistemic adverbs like \textit{maybe} \citep{oneill2000maybe}, suggesting that conceptual capacity for modal notions may be available earlier.

Another account focuses on input frequency. The distribution of modal senses in child-directed English differs from what is attested in non-CDL corpora, and researchers have argued that this discrepancy may influence how children acquire modals. Accordingly, \citet{vandooren2022figuring} show that the epistemic modal sense, which emerges late in children's acquisition, is highly infrequent in children's input.

The frequencies of the different modal senses have also been shown to vary depending on the specific modal. In American English corpora consisting mostly of written text, \textit{may} appears as epistemic 83\% of the time and as root only 17\% of the time \citep{collins2007}. For \textit{must}, the pattern reverses, with root uses accounting for 84\% of occurrences and epistemic uses only 16\% \citep{hw2012}. The Manchester corpus \citep{theakston2001role}, as sample of CDL corpora, reveals a contrasting distribution. In this case, \textit{may} occurs as root 56\% of the time and as epistemic 44\% of the time, whereas \textit{must} occurs as epistemic 65\% of the time and as root 35\% of the time \citep{vandooren2022figuring}. These diverging distributions also provide a basis for testing whether LMs trained on different input types develop corresponding frequency-based preferences.

\section{Evaluation Data}

\begin{figure}
    \includegraphics[width=0.55\textwidth]{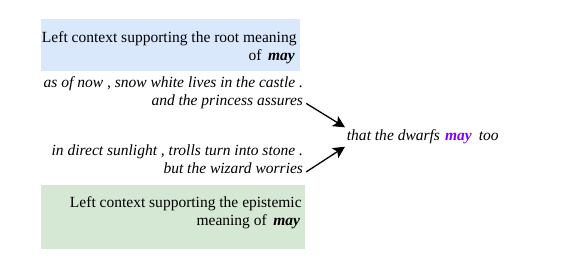}
    \caption{Example data pair from the validated dataset.}
    \label{fig:val_example}
\end{figure}

There are no standardized benchmarks for modal sense disambiguation. For evaluation purposes, we consider four datasets, which we refer to as EPOS, validated, naturalistic, and naturalistic with context. The datasets differ in their source material, size, and the availability of disambiguating context, making them complementary to each other. The EPOS dataset is the largest of all. The validated dataset originates from a highly controlled neurolinguistic experiment, allowing us to use carefully constructed stimuli in pairwise comparisons. The naturalistic dataset, drawn from an annotated CDL corpus, lets us examine how results differ when LMs are presented the kind of disambiguation tasks that children face. Finally, its variant with retrieved prior context further allows us to assess whether the addition of naturally occurring context affects LM performance. Example data points from the validated dataset can be seen in Figure~\ref{fig:val_example}. Appendix \ref{sec:data_example} provides further examples for each dataset, along with the number of instances per modal class.

\subsection{EPOS}
The publicly available EPOS dataset \citep{zhou-etal-2015-semantically} is constructed through paraphrase-driven modal sense projection on the Europarl \citep{koehn-2005-europarl} and OpenSubtitles corpora \citep{tiedemann-2012-parallel}, which feature European Parliament proceedings and film subtitles, respectively. Rather than relying on manual annotation, the authors assign labels heuristically using paraphrase patterns. While this approach introduces some noise, it also provides a means of obtaining a sizable corpus for modal sense classification. The dataset exhibits some imbalance, containing 134 root and 875 epistemic \textit{may} instances, alongside 417 root and 637 epistemic instances of \textit{must}. Several existing studies on modal sense classification have used this dataset \citep{marasovic2016multilingual, li2019modal, dehouck2023revisiting, wagner2023probing}.

\subsection{Human-validated}
This dataset constructed by \citet{tulling2020neural} for their neurolinguistic experiments consists of 160 stimuli: 40 instances each of \textit{must} in a root context, \textit{must} in an epistemic context, \textit{may} in a root context, and \textit{may} in an epistemic context. Twenty-five adult participants (aged 19--52) validated each sentence to be associated with the intended modal sense. Each test item consists of a context sentence supporting either an epistemic or root interpretation, followed by the target sequence containing the modal (see example in Figure~\ref{fig:val_example}). The target sequence features an embedded clause within a matrix clause linking the modal's interpretation to a third-person perspective to further bias the intended reading. Importantly, the embedded clause itself (introduced by ``that'') is kept identical across root and epistemic conditions, while the matrix verb varies. The target sequence is identical across each root-epistemic pair, allowing for a more controlled evaluation setting. 

\subsection{Naturalistic}
The naturalistic data are drawn from an annotated subset of the CHILDES database \citep{macwhinney2000childes}. \Citet{vandooren2022figuring} extract modal productions from the Manchester corpus \citep{theakston2001role}, which consists of recordings from 12 child--mother dyads during play sessions (ages 2;00 to 3;00), and annotate each instance for root or epistemic sense. From this dataset, we use 296 instances of \textit{must} (88 root, 208 epistemic) and 32 instances of \textit{may} (16 root, 16 epistemic), restricting our set to the utterances we could locate in the database.

We additionally create a version of this dataset with conversational context. We utilize childes-db \citep[version 2018.1;][]{sanchez2019childes} to retrieve from the transcripts the preceding utterances for each target sentence. In each case, we extract up to five prior adult utterances from the same transcript file. The average utterance length in the corpus is 3.89 words, so retrieving five previous utterances yields a context window of approximately 20 words. When fewer than five prior adult utterances are available, we include all that are present.

\section{Methods}

We evaluate seven pretrained LMs that vary in architecture, size, and training data type. We first test each LM's ability to discriminate between epistemic and root senses of \textit{may} and \textit{must} using vector-based similarity probing. We then analyze LM preferences for one modal sense over the other using surprisal-based measures in a test design analogous to the standard minimal-pair setup.

\subsection{Probing vector representations}
\label{sec:method-similarity}

We adopt the word sense disambiguation method of \citet{cabiddu2023comparing}, who use sense prototypes derived from contextualized embeddings to disambiguate polysemous words (e.g., distinguishing the elastic band meaning of \textit{band} from the music group meaning) and adapt it to our use case of distinguishing epistemic and root modal senses.  We divide each of our datasets into a prototype set and an evaluation set using stratified 5-fold cross-validation. For every combination of modal (\textit{must}, \textit{may}) and sense (epistemic, root), we extract prototypical vector representations from the prototype set. Given an occurrence of the modal in a particular sense, we extract its contextualized representation from the model and sum the representations from the last four hidden layers, following \citet{cabiddu2023comparing} and \citet{loureiro-etal-2021-analysis}. These vectors are then averaged across training instances for each sense to obtain a centroid.

During evaluation, we again extract the contextualized vector for each test instance of the target modal by summing the last four hidden layers. We compute the cosine similarity between this test vector and the two sense prototypes (epistemic and root). The predicted sense is taken to be the prototype with higher similarity. Accuracy is computed as the percentage of test instances where the predicted sense matches the ground-truth label. As a shorthand, we refer to tests performed using this method as the \textit{similarity method}.

Intuitively, this method differs from a typical probing classifier setup \citep{ettinger-etal-2016-probing, alain2018understandingintermediatelayersusing} in that it directly compares distances between vectors in a model's representation space.

\subsection{Surprisal-based measures}
\label{sec:method-surprisal}

Surprisal is a standard measure for estimating the likelihood of a stimulus under a given model. We compute token-level surprisal as the negative log probability of a token given its preceding context. For masked LMs, we use pseudo-surprisal scores obtained by successively masking tokens following the widely used method by \citet{salazar-etal-2020-masked}. Lower surprisal values indicate that a model finds a token more predictable in its context, and higher values indicate that the token is less predictable.

Similar to a standard minimal-pair setup, we use this measure to evaluate sentence pairs where the target sequences are identical and the preceding context disambiguates the intended sense (see Figure~\ref{fig:val_example}). For each model, we compute the surprisal of the modal token in both the root and epistemic context conditions. A lower surprisal value for one condition relative to the other is interpreted as a preference for that modal sense. No subword aggregation was required for our surprisal estimates, as none of the LMs split the modal tokens across multiple pieces. Because this method requires paired data points that differ only by a root versus epistemic context while keeping the target sequence identical, we perform surprisal-based evaluation only on the validated dataset.

\begin{table*}[ht]
\centering
\small
\renewcommand{\arraystretch}{1.1}
\setlength{\tabcolsep}{4pt}
\begin{tabular}{@{}l c l l *{6}{>{\centering\arraybackslash}p{1.25cm}}@{}}
 & & & & \multicolumn{3}{c}{May} & \multicolumn{3}{c}{Must} \\
\cmidrule(lr){5-7} \cmidrule(lr){8-10}
 & & & & Root & Epistemic & All & Root & Epistemic & All \\
\hline \noalign{\vspace{0.3ex}}
\multirow[l]{12}{*}{\rotatebox{90}{Masked LMs}} & & \multirow[l]{4}{*}{\shortstack[l]{RoBERTa\\\texttt{(110M)}}} & EPOS & \cellcolor[rgb]{0.45,0.73,0.43} 0.95 & \cellcolor[rgb]{0.40,0.71,0.40} 0.99 & \cellcolor[rgb]{0.41,0.71,0.40} 0.98 & \cellcolor[rgb]{0.45,0.73,0.43} 0.95 & \cellcolor[rgb]{0.55,0.78,0.49} 0.87 & \cellcolor[rgb]{0.51,0.76,0.47} 0.90 \\
 & & & Validated & \cellcolor[rgb]{0.39,0.71,0.39} 1.00 & \cellcolor[rgb]{0.39,0.71,0.39} 1.00 & \cellcolor[rgb]{0.39,0.71,0.39} 1.00 & \cellcolor[rgb]{0.48,0.75,0.45} 0.93 & \cellcolor[rgb]{0.45,0.73,0.43} 0.95 & \cellcolor[rgb]{0.47,0.74,0.44} 0.94 \\
 & & & Naturalistic & \cellcolor[rgb]{0.73,0.87,0.61} 0.72 & \cellcolor[rgb]{0.39,0.71,0.39} 1.00 & \cellcolor[rgb]{0.58,0.79,0.51} 0.85 & \cellcolor[rgb]{0.55,0.78,0.50} 0.86 & \cellcolor[rgb]{0.56,0.78,0.50} 0.86 & \cellcolor[rgb]{0.56,0.78,0.50} 0.86 \\
 & & & Naturalistic (with context) & \cellcolor[rgb]{0.80,0.90,0.65} 0.67 & \cellcolor[rgb]{0.39,0.71,0.39} 1.00 & \cellcolor[rgb]{0.61,0.81,0.53} 0.82 & \cellcolor[rgb]{0.53,0.77,0.48} 0.89 & \cellcolor[rgb]{0.44,0.73,0.42} 0.96 & \cellcolor[rgb]{0.47,0.74,0.44} 0.94 \\
\cline{3-10}\noalign{\vspace{0.4ex}}
 & & \multirow[l]{4}{*}{\shortstack[l]{CHILDES\\RoBERTa\\\texttt{(12.7M)}}} & EPOS & \cellcolor[rgb]{0.50,0.76,0.46} 0.91 & \cellcolor[rgb]{0.56,0.79,0.50} 0.86 & \cellcolor[rgb]{0.56,0.78,0.50} 0.86 & \cellcolor[rgb]{0.75,0.88,0.62} 0.70 & \cellcolor[rgb]{0.63,0.82,0.55} 0.80 & \cellcolor[rgb]{0.68,0.84,0.58} 0.76 \\
 & & & Validated & \cellcolor[rgb]{1,0.85,0.67} 0.42 & \cellcolor[rgb]{0.94,0.97,0.74} 0.55 & \cellcolor[rgb]{1,0.97,0.76} 0.49 & \cellcolor[rgb]{0.94,0.97,0.74} 0.55 & \cellcolor[rgb]{0.94,0.97,0.74} 0.55 & \cellcolor[rgb]{0.94,0.97,0.74} 0.55 \\
 & & & Naturalistic & \cellcolor[rgb]{0.59,0.80,0.52} 0.83 & \cellcolor[rgb]{0.63,0.82,0.55} 0.80 & \cellcolor[rgb]{0.61,0.81,0.53} 0.82 & \cellcolor[rgb]{0.62,0.82,0.54} 0.81 & \cellcolor[rgb]{0.59,0.80,0.52} 0.84 & \cellcolor[rgb]{0.60,0.80,0.53} 0.83 \\
 & & & Naturalistic (with context) & \cellcolor[rgb]{0.67,0.84,0.57} 0.77 & \cellcolor[rgb]{0.39,0.71,0.39} 1.00 & \cellcolor[rgb]{0.54,0.78,0.49} 0.88 & \cellcolor[rgb]{0.60,0.80,0.52} 0.83 & \cellcolor[rgb]{0.48,0.75,0.45} 0.92 & \cellcolor[rgb]{0.52,0.76,0.47} 0.90 \\
\cline{3-10}\noalign{\vspace{0.4ex}}
 & & \multirow[l]{4}{*}{\shortstack[l]{Wikipedia\\RoBERTa\\\texttt{(12.7M)}}} & EPOS & \cellcolor[rgb]{0.57,0.79,0.51} 0.85 & \cellcolor[rgb]{0.45,0.73,0.43} 0.95 & \cellcolor[rgb]{0.47,0.74,0.44} 0.94 & \cellcolor[rgb]{0.78,0.89,0.64} 0.68 & \cellcolor[rgb]{0.59,0.80,0.52} 0.84 & \cellcolor[rgb]{0.67,0.84,0.57} 0.77 \\
 & & & Validated & \cellcolor[rgb]{0.88,0.94,0.71} 0.60 & \cellcolor[rgb]{1,0.55,0.43} 0.28 & \cellcolor[rgb]{1,0.87,0.69} 0.44 & \cellcolor[rgb]{1,0.95,0.75} 0.47 & \cellcolor[rgb]{1,0.55,0.43} 0.28 & \cellcolor[rgb]{1,0.75,0.59} 0.38 \\
 & & & Naturalistic & \cellcolor[rgb]{0.73,0.87,0.61} 0.72 & \cellcolor[rgb]{0.63,0.82,0.55} 0.80 & \cellcolor[rgb]{0.69,0.85,0.58} 0.76 & \cellcolor[rgb]{0.60,0.80,0.52} 0.83 & \cellcolor[rgb]{0.50,0.76,0.46} 0.91 & \cellcolor[rgb]{0.53,0.77,0.48} 0.89 \\
 & & & Naturalistic (with context) & \cellcolor[rgb]{0.73,0.87,0.61} 0.72 & \cellcolor[rgb]{0.47,0.74,0.44} 0.93 & \cellcolor[rgb]{0.62,0.81,0.54} 0.81 & \cellcolor[rgb]{0.60,0.80,0.52} 0.83 & \cellcolor[rgb]{0.46,0.74,0.44} 0.94 & \cellcolor[rgb]{0.50,0.76,0.46} 0.91 \\
\hline \noalign{\vspace{0.3ex}}
\multirow[l]{16}{*}{\rotatebox{90}{Causal LMs}} & & \multirow[l]{4}{*}{\shortstack[l]{Pythia\\\texttt{(6.9B)}}} & EPOS & \cellcolor[rgb]{1,0.53,0.42} 0.27 & \cellcolor[rgb]{0.40,0.71,0.40} 0.99 & \cellcolor[rgb]{0.52,0.77,0.47} 0.89 & \cellcolor[rgb]{0.41,0.71,0.40} 0.98 & \cellcolor[rgb]{1,0.38,0.30} 0.19 & \cellcolor[rgb]{0.99,1.00,0.78} 0.50 \\
 & & & Validated & \cellcolor[rgb]{0.51,0.76,0.47} 0.90 & \cellcolor[rgb]{0.39,0.71,0.39} 1.00 & \cellcolor[rgb]{0.45,0.73,0.43} 0.95 & \cellcolor[rgb]{0.48,0.75,0.45} 0.93 & \cellcolor[rgb]{0.48,0.75,0.45} 0.93 & \cellcolor[rgb]{0.48,0.75,0.45} 0.93 \\
 & & & Naturalistic & \cellcolor[rgb]{0.96,0.98,0.76} 0.53 & \cellcolor[rgb]{0.69,0.85,0.59} 0.75 & \cellcolor[rgb]{0.83,0.92,0.67} 0.64 & \cellcolor[rgb]{0.83,0.92,0.67} 0.64 & \cellcolor[rgb]{0.90,0.95,0.72} 0.58 & \cellcolor[rgb]{0.88,0.94,0.71} 0.60 \\
 & & & Naturalistic (with context) & \cellcolor[rgb]{0.73,0.87,0.61} 0.72 & \cellcolor[rgb]{0.63,0.82,0.55} 0.80 & \cellcolor[rgb]{0.69,0.85,0.58} 0.76 & \cellcolor[rgb]{0.64,0.83,0.55} 0.79 & \cellcolor[rgb]{0.67,0.84,0.57} 0.77 & \cellcolor[rgb]{0.66,0.83,0.56} 0.78 \\
\cline{3-10} \noalign{\vspace{0.4ex}}
 & & \multirow[l]{4}{*}{\shortstack[l]{GPT-2\\\texttt{(124M)}}} & EPOS & \cellcolor[rgb]{1,0.52,0.41} 0.26 & \cellcolor[rgb]{0.40,0.71,0.40} 0.99 & \cellcolor[rgb]{0.52,0.77,0.47} 0.89 & \cellcolor[rgb]{0.41,0.71,0.40} 0.99 & \cellcolor[rgb]{1,0.16,0.12} 0.08 & \cellcolor[rgb]{1,0.87,0.68} 0.44 \\
 & & & Validated & \cellcolor[rgb]{0.42,0.72,0.41} 0.97 & \cellcolor[rgb]{0.45,0.73,0.43} 0.95 & \cellcolor[rgb]{0.44,0.73,0.42} 0.96 & \cellcolor[rgb]{0.42,0.72,0.41} 0.97 & \cellcolor[rgb]{0.42,0.72,0.41} 0.97 & \cellcolor[rgb]{0.42,0.72,0.41} 0.97 \\
 & & & Naturalistic & \cellcolor[rgb]{0.88,0.94,0.71} 0.60 & \cellcolor[rgb]{1,0.53,0.42} 0.27 & \cellcolor[rgb]{1,0.84,0.65} 0.42 & \cellcolor[rgb]{0.64,0.82,0.55} 0.80 & \cellcolor[rgb]{1,0.51,0.40} 0.26 & \cellcolor[rgb]{1,0.83,0.65} 0.42 \\
 & & & Naturalistic (with context) & \cellcolor[rgb]{0.71,0.86,0.60} 0.73 & \cellcolor[rgb]{0.92,0.96,0.73} 0.57 & \cellcolor[rgb]{0.83,0.92,0.67} 0.64 & \cellcolor[rgb]{0.77,0.89,0.63} 0.69 & \cellcolor[rgb]{0.69,0.85,0.59} 0.75 & \cellcolor[rgb]{0.71,0.86,0.60} 0.73 \\
\cline{3-10} \noalign{\vspace{0.4ex}}
 & & \multirow[l]{4}{*}{\shortstack[l]{CHILDES\\GPT-2\\\texttt{(14.8M)}}} & EPOS & \cellcolor[rgb]{0.99,0.99,0.78} 0.51 & \cellcolor[rgb]{0.47,0.75,0.44} 0.93 & \cellcolor[rgb]{0.54,0.78,0.49} 0.88 & \cellcolor[rgb]{0.64,0.83,0.55} 0.79 & \cellcolor[rgb]{0.70,0.85,0.59} 0.74 & \cellcolor[rgb]{0.68,0.84,0.58} 0.76 \\
 & & & Validated & \cellcolor[rgb]{0.76,0.88,0.63} 0.70 & \cellcolor[rgb]{0.78,0.89,0.64} 0.68 & \cellcolor[rgb]{0.77,0.89,0.64} 0.69 & \cellcolor[rgb]{0.78,0.89,0.64} 0.68 & \cellcolor[rgb]{0.88,0.94,0.71} 0.60 & \cellcolor[rgb]{0.83,0.92,0.67} 0.64 \\
 & & & Naturalistic & \cellcolor[rgb]{0.71,0.86,0.60} 0.73 & \cellcolor[rgb]{0.78,0.89,0.64} 0.68 & \cellcolor[rgb]{0.75,0.88,0.62} 0.70 & \cellcolor[rgb]{0.79,0.90,0.65} 0.67 & \cellcolor[rgb]{0.66,0.84,0.57} 0.77 & \cellcolor[rgb]{0.70,0.85,0.59} 0.74 \\
 & & & Naturalistic (with context) & \cellcolor[rgb]{0.65,0.83,0.56} 0.78 & \cellcolor[rgb]{0.69,0.85,0.59} 0.75 & \cellcolor[rgb]{0.68,0.84,0.58} 0.76 & \cellcolor[rgb]{0.68,0.84,0.58} 0.76 & \cellcolor[rgb]{0.74,0.87,0.62} 0.71 & \cellcolor[rgb]{0.72,0.87,0.60} 0.73 \\
\cline{3-10} \noalign{\vspace{0.4ex}}
 & & \multirow[l]{4}{*}{\shortstack[l]{Wikipedia\\GPT-2\\\texttt{(14.8M)}}} & EPOS & \cellcolor[rgb]{0.88,0.94,0.71} 0.60 & \cellcolor[rgb]{0.42,0.72,0.41} 0.98 & \cellcolor[rgb]{0.48,0.75,0.45} 0.93 & \cellcolor[rgb]{0.67,0.84,0.57} 0.77 & \cellcolor[rgb]{0.68,0.85,0.58} 0.76 & \cellcolor[rgb]{0.68,0.84,0.58} 0.76 \\
 & & & Validated & \cellcolor[rgb]{0.69,0.85,0.59} 0.75 & \cellcolor[rgb]{0.73,0.87,0.61} 0.72 & \cellcolor[rgb]{0.71,0.86,0.60} 0.74 & \cellcolor[rgb]{0.76,0.88,0.63} 0.70 & \cellcolor[rgb]{0.85,0.93,0.69} 0.62 & \cellcolor[rgb]{0.80,0.90,0.65} 0.66 \\
 & & & Naturalistic & \cellcolor[rgb]{0.82,0.91,0.67} 0.65 & \cellcolor[rgb]{0.76,0.88,0.63} 0.70 & \cellcolor[rgb]{0.79,0.90,0.65} 0.67 & \cellcolor[rgb]{0.72,0.87,0.60} 0.73 & \cellcolor[rgb]{0.71,0.85,0.59} 0.74 & \cellcolor[rgb]{0.71,0.86,0.60} 0.74 \\
 & & & Naturalistic (with context) & \cellcolor[rgb]{0.90,0.95,0.72} 0.58 & \cellcolor[rgb]{0.92,0.96,0.73} 0.57 & \cellcolor[rgb]{0.91,0.96,0.73} 0.57 & \cellcolor[rgb]{0.85,0.93,0.69} 0.62 & \cellcolor[rgb]{0.84,0.92,0.68} 0.63 & \cellcolor[rgb]{0.84,0.92,0.68} 0.63 \\
\hline
\end{tabular}
\caption{Modal sense disambiguation accuracy by different LMs, based on the vector similarity method (cf. Section \ref{sec:method-similarity}). The red-green color gradient represents performance from low (red) to high (green), with the 0.50 midpoint marking random chance levels. Parameter counts are indicated in indicated in parentheses for each LM. See Appendix \ref{sec:sim_plots} for more detailed representations of the results.}
\label{tab:similarity_heatmap}
\end{table*}

\subsection{Models}

We evaluate a total of seven LMs with different architectures, sizes, and training data sources. Our first group of LMs consists of base models pretrained on relatively large amounts of data. Pythia 6.9B \citep{biderman2023pythia} is the largest LM we include, belonging to a suite of decoder-only open-source models whose training data are publicly available. We also include GPT-2 \citep{radford2019language}, an autoregressive decoder-only model, and RoBERTa \citep{liu2019roberta}, an optimized bidirectional encoder-only model. We use the pretrained base versions for all three LMs.

To provide a more developmentally plausible reference point, we also employ LMs released by \citet{padovani2025child} which have been trained exclusively on CDL. These are RoBERTa and GPT-2 variants trained from scratch on the AO-CHILDES corpus \citep{huebner2021lex}, which contains approximately 5 million words of American English CDL. We refer to these as CHILDES RoBERTa and CHILDES GPT-2. Given that the distribution of modal senses can differ for CDL compared to general text, we include these LMs because they are trained exclusively on CDL without additional data sources.

We also use RoBERTa and GPT-2 models trained on a Wikipedia corpus matching the size of the CDL data, also released by \citet{padovani2025child}. Testing these Wikipedia RoBERTa and Wikipedia GPT-2 models allows us to isolate the effect of training data type (CDL vs.\ general text) from the effect of corpus size. No fine-tuning is performed on any of the listed LMs.

\section{Results and Discussion}

\subsection{Use of additional context}

The first question we address is whether additional discourse context helps LMs distinguish between root and epistemic meanings of modals. We expect LMs to perform better when evaluated on datasets that include prior context, under the assumption that the preceding sentence can help disambiguate the modal in the target utterance. We further expect the validated dataset to be the easiest among all evaluation sets, given that it features context sentences carefully designed to bias people toward a particular modal reading. In contrast, the EPOS dataset and the naturalistic dataset without context present the modal sentence in isolation without any prior context to guide interpretation. It should be noted that for the naturalistic dataset with context, we simply retrieve preceding utterances to provide context. Unlike the carefully designed stimuli in the validated dataset, these may not necessarily contain the relevant information to disambiguate modal senses.

Based on the results in Table \ref{tab:similarity_heatmap}, the availability of adequate prior context is generally associated with improved performance. LMs tend to perform substantially better on the validated dataset than on the EPOS dataset or the naturalistic dataset without context, with most LMs achieving accuracies above 90\%. That being said, this pattern only holds for the larger LMs. Our set of developmentally plausible LMs exhibit near-chance results on the validated dataset, suggesting that they may not have learned modal representations sufficiently robust to take advantage of the disambiguating context.

To investigate how context affects results, we also compare model performance on the naturalistic dataset with and without prior context. For GPT-2, \textit{may} epistemic improves from 27\% to 57\%, and for Pythia from 75\% to 80\%. Likewise, \textit{must} epistemic shows an improvement from 26\% to 75\% for GPT-2 and from 58\% to 77\% for Pythia. We observe that the inclusion of contextual cues improves performance for causal LMs. Masked LMs similarly benefit from additional left context, although the improvement is less marked. Notably, RoBERTa shows \textit{must} epistemic results improving from 86\% to 96\%, while \textit{may} root decreases from 72\% to 67\%. Appendix \ref{sec:validated_variations} offers a supplementary investigation into the usefulness of different contextual cues.

Looking at these results, we can additionally note an architectural difference. Causal LMs show lower performance on datasets that present greater difficulty for modal sense disambiguation, namely the EPOS dataset and the naturalistic data without context. Masked LMs, in contrast, show more uniform performance across all datasets. We attribute this difference to the bidirectional context available to masked LMs. \Citet{vandooren2022figuring}, building on insights from \citet{Condoravdi2002}, hypothesize that aspectual cues like \textit{must be} or \textit{may have} (where the words \textit{be} or \textit{have} appear in the right context of the target modal) can be particularly helpful when disambiguating modal senses. 
Having the ability to leverage such information could explain the difference in model behaviors between causal and masked LMs.

\subsection{Effect of out-of-domain data}

We further examine whether LMs trained on data from the same domain as the test data show better ability to discriminate between modal senses. Our domain of interest here is CDL, and we expect the evaluation on naturalistic data (which originates from CHILDES) to be more challenging for LMs not trained on CDL, leading to lower performance and more varied behavior. Conversely, we expect CDL-trained LMs to perform better on the naturalistic dataset than other evaluation sets. Since the EPOS dataset does not have prior context available, we are interested in a comparison with no-context variant of the naturalistic dataset.

As Table \ref{tab:similarity_heatmap} illustrates, we observe that GPT-2 scores as low as 27\% for epistemic \textit{may} and 26\% for epistemic \textit{must} on the naturalistic dataset. This low performance aligns with the expectation that out-of-domain data poses challenges for LMs not trained on CDL. Pythia also shows lower performance on naturalistic datasets, with scores between 53\% and 80\%. For RoBERTa, the naturalistic dataset likewise proves slightly more difficult than EPOS or the validated dataset.

CDL-trained LMs, however, do not exhibit the expected in-domain benefit. CHILDES GPT-2 achieves its highest overall scores on the EPOS dataset, with accuracies of 88\% for \textit{may} and 76\% for \textit{must}, compared to 70\% and 74\% on the naturalistic dataset. CHILDES RoBERTa performs slightly better on EPOS for \textit{may}, scoring 86\% versus 82\% on naturalistic data, while the pattern reverses for \textit{must}, with 76\% on EPOS against 83\% on naturalistic data. Notably, both CDL- and Wikipedia-trained LMs produce similar results to each other while diverging from larger LMs like RoBERTa, GPT-2, and Pythia. This suggests that model size, rather than training data domain, is a larger contributor to the observed differences between the smaller developmentally plausible LMs and those that are larger.

\begin{figure*}[ht]
    \centering
    \includegraphics[]{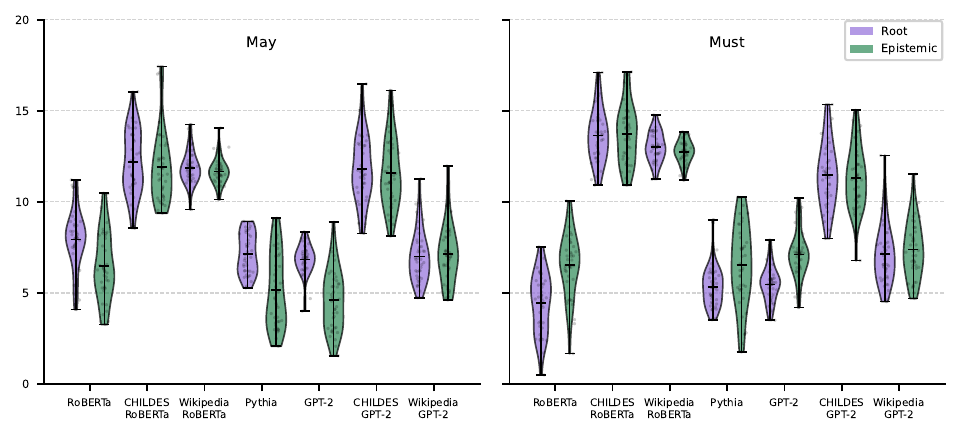}
    \caption{Surprisal score distributions for the modals in validated data. Lower surprisal values for a given sense are interpreted as a preference for that sense.}
    \label{fig:validated_surprisal}
\end{figure*}

\subsection{Role of input frequency}

Finally, we investigate whether LMs' ability to distinguish between root and epistemic uses of \textit{may} and \textit{must} reflects the distribution of these senses in their training data. Here, we consider both the vector-probing and surprisal-based evaluation. If LM behavior aligns with input frequency, LMs should perform better on the more frequent sense of each modal under the similarity method and show a preference for that sense under the surprisal method.

For LMs trained on general text, we assume their training data matches attested corpus statistics, where \textit{may} is predominantly epistemic \citep[83\%;][]{collins2007} and \textit{must} is predominantly root \citep[84\%;][]{hw2012}. For CDL-trained LMs, we extrapolate from the Manchester corpus, where \textit{may} is root 56\% of the time and \textit{must} is epistemic 65\% of the time \citep{vandooren2022figuring}. For Wikipedia-trained LMs, we assume their training data aligns more closely with general corpus statistics, as Wikipedia consists of written encyclopedic text. Therefore, we expect non-CDL-trained LMs to perform better on epistemic \textit{may} and root \textit{must}, whereas CDL-trained LMs should perform better on root \textit{may} and epistemic \textit{must}. A breakdown of the corpus statistics and model-specific expectations can be found in Appendix \ref{sec:corpus_stats}. We further reflect on our corpus-based assumptions in the limitations section.

Similarity method results in Table \ref{tab:similarity_heatmap} mostly follow the expected pattern for non-CDL-trained LMs, particularly when evaluating on harder-to-disambiguate data. As noted earlier, the EPOS dataset and the naturalistic dataset without context are more likely to contain ambiguous modals, as they present the modal sentence in isolation without any preceding context. Lacking the discourse context needed to disambiguate the intended reading, evaluation on these datasets can uncover whether LMs rely on any internal biases when it comes to modal senses.

Looking at the EPOS dataset, causal LMs trained on general text can be said to show a clear alignment with input frequency. GPT-2 achieves near-ceiling performance on the dominant senses (99\% for epistemic \textit{may} and 99\% for root \textit{must}) while scoring at or below chance on the infrequent senses (26\% for root \textit{may} and 8\% for epistemic \textit{must}). Pythia follows a similar pattern, scoring 99\% for epistemic \textit{may} and 98\% for root \textit{must} versus 27\% and 19\% for the infrequent senses, respectively. Masked LMs trained on general text, specifically RoBERTa, likewise align with the expected pattern in most cases, though their results are very uniform across both senses, resulting in higher overall performance.
This uniformity suggests that input frequency may have a less prominent effect on the behavior of masked LMs.

Turning to the validated dataset, we can observe that non-CDL-trained LMs do not always show the expected pattern. GPT-2 scores 97\% for root \textit{may}, 95\% for epistemic \textit{may}, and 97\% for both senses of \textit{must}. This indicates that when evaluation data includes helpful disambiguating context, LMs can take advantage of it. The correlation with input frequency can be observed only when such context is absent.

Two exceptions to this pattern are worth noting. First, on the naturalistic dataset without context, GPT-2 does not maintain the expected pattern for \textit{may}. Performance on epistemic \textit{may} drops to 27\%, though performance on root \textit{must} remains strong at 80\%. This variability may stem from naturalistic data being out of domain for GPT-2, or from low \textit{may} counts in the evaluation data producing variable outcomes. Second, Wikipedia-trained LMs do not consistently exhibit the expected pattern despite being trained on general text. Their scores more closely resemble those of CDL-trained LMs than those of larger LMs trained on general text.

For LMs trained on CDL, we do not consistently observe the opposite pattern that we expect based on the Manchester corpus statistics. On the EPOS dataset, CHILDES GPT-2 scores 51\% for root \textit{may}, 93\% for epistemic \textit{may}, 79\% for root \textit{must}, and 74\% for epistemic \textit{must}, while CHILDES RoBERTa scores 91\%, 86\%, 70\%, and 80\%, respectively. On the validated dataset, developmentally plausible LMs tend to perform near chance levels for the different modal senses. We can therefore note that the developmentally plausible LMs are not consistently better at one modal sense over the other on any dataset, suggesting that they may not have encountered enough modal examples in training to develop a sensitivity in either direction.

To explore whether frequency also affects LMs' preferences, we also consider the surprisal-based results in Figure \ref{fig:validated_surprisal}. Consistent with our similarity-based results, we can note that the surprisal distributions on the validated dataset differ between larger LMs and developmentally plausible ones. Non-CDL-trained larger LMs show the expected preference for epistemic \textit{may} and root \textit{must}. Both CDL-trained and Wikipedia-trained smaller LMs lack a clear preference for either sense of \textit{must} or \textit{may}, with average surprisal values remaining roughly equal across the two senses. Given that our developmentally plausible LMs have very similar surprisal distributions for both epistemic and root modals, we can argue that these smaller LMs have not been able to develop distinct representations for different modal senses, consistent with our similarity-based results.

\section{Conclusion}

This study investigates whether LMs can distinguish between root and epistemic senses of the English modals \textit{may} and \textit{must}. We ask whether providing additional conversational context helps LMs discriminate between modal senses, whether LMs trained on data from the same domain as evaluation data show better discrimination ability, and whether LM behavior correlates with the estimated distribution of modal senses in their training data. We evaluate seven LMs that vary in architecture, size, and training data type using vector-based similarity probing and surprisal measures across different datasets.

We show that providing additional context substantially improves the performance of larger LMs, whereas performance is lower when they are evaluated on ambiguous modal expressions without contextual cues to resolve ambiguity. This suggests that larger LMs are capable of leveraging contextual information to override existing biases if such information is present.

Furthermore, we find that LM size affects modal sense discrimination more than training domain. Smaller developmentally plausible LMs (both CDL-trained and Wikipedia-trained) perform worse than larger LMs. Critically, if training domain were the key factor, Wikipedia-trained LMs would pattern with larger LMs trained on general text and not with their CDL-trained counterparts, as we observe.

Looking at whether LMs' performance aligns with input frequencies, we observe that larger LMs trained on general text perform better on the dominant sense of each modal under the similarity method and show a preference for that sense under the surprisal method. This alignment with corpus statistics only emerges when evaluation data lack disambiguating context, indicating that LMs can rely on general biases when contextual cues are absent.

Overall, our results offer insights into how LMs represent modal senses. We see that sufficiently large LMs can disambiguate modals through frequency biases or contextual cues, whereas smaller LMs trained on limited input fail to acquire reliable sense distinctions. More broadly, our findings bridge computational modeling and linguistic theories of modal acquisition by showing that two factors hypothesized to implicate children's learning, namely input frequency and the integration of contextual information, also shape LM behavior.

\section*{Limitations}

While our study provides a systematic investigation into how LMs distinguish between root and epistemic senses of \textit{may} and \textit{must}, several limitations can be acknowledged. First, we can note that our evaluations target only two English modals, \textit{may} and \textit{must}. We, therefore, cannot reliably comment on whether the observed patterns generalize to other modals such as \textit{can}, \textit{should}, and \textit{could}, or to other languages.

The largest LM we assess is Pythia with 6.9 billion parameters. Whether larger LMs would behave differently, for instance by being able to disambiguate modal senses without rich contextual cues, remains an open question. Consequently, our conclusions about model size apply only within the parameter range we investigate.

Although we argue that our smaller developmentally plausible LMs may have low performance due to insufficient exposure to modal examples, this interpretation is inferential. We do not know the exact number of modal expressions that the various LMs have been exposed to. Consequently, our analyses rely on existing corpus studies rather than direct measurement to characterize frequency distributions in each LM's training data. For models with publicly available data, raw token counts of \textit{may} and \textit{must} would offer no insight into root versus epistemic distributions. Extracting sense-level frequencies would require annotating modal senses for sizeable datasets, which is a prohibitively large annotation effort. We can also note that any automated annotation would have to depend on the LMs we are evaluating. Some degree of extrapolation is therefore inevitable, and our choice of frequency proxies is grounded in existing linguistic literature and constrained by the availability of annotated datasets. Nevertheless, we acknowledge this concern and emphasize that our frequency-based arguments should be interpreted in light of this limitation.

Finally, our evaluation introduces a potential language variety confound. The CDL models are trained exclusively on American English, whereas the Manchester corpus is British English. We utilize this corpus for evaluations with naturalistic data and also for the modal sense statistics that guide our expectations about CDL-trained LM behavior. While it would have been preferable to have American English CDL frequency statistics for comparison, the only available CDL dataset annotated with modal sense is the Manchester corpus, which restricts us to this variety. However, this is not a confounder when comparing CDL-trained models to other LMs, as they are also predominantly trained on American English. We acknowledge this limitation while noting that \citet{collins2007} reports similar frequencies for root and epistemic \textit{may} in adult American English and in British English, suggesting that the overall patterns may still hold.

\section*{Acknowledgments}

We thank the anonymous reviewers for their valuable feedback. This publication was supported by the Talent Programme of the Dutch Research Council (grant VI.Vidi.221C.009).

\bibliography{custom}

\appendix

\onecolumn

\section{Dataset details}

\label{sec:data_example}

\begin{table}[ht]
\centering
\small
\renewcommand{\arraystretch}{1.4}
\setlength{\tabcolsep}{4pt}
\begin{tabular}{|>{\centering\arraybackslash}m{0.7cm}|>{\raggedright\arraybackslash}p{1.3cm}|p{5.8cm}|p{5.75cm}|r|}
\hline
Data & Flavor & Context & Target sequence & \# \\
\hline
\hline
\multirow{4}{*}{\centering \rotatebox[origin=c]{90}{\centering EPOS \phantom{000}}} & \raggedright Root & \cellcolor[HTML]{E8E8E8}{} & i will now , if i \textbf{\color[HTML]{5B3A7A}{\underline{may}}} , read out the key points of the recommendations we adopted today . & 134 \\
\cline{2-5}
& \raggedright Epistemic & \cellcolor[HTML]{E8E8E8}{} & this wording \textbf{\color[HTML]{3D6B4F}{\underline{may}}} sound a little far-fetched and need some explanation . & 875 \\
\cline{2-5}
& \raggedright Root & \cellcolor[HTML]{E8E8E8}{} & we \textbf{\color[HTML]{5B3A7A}{\underline{must}}} be prepared for a long haul . & 417 \\
\cline{2-5}
& \raggedright Epistemic & \cellcolor[HTML]{E8E8E8}{} & there \textbf{\color[HTML]{3D6B4F}{\underline{must}}} be a lower back injury as well . & 637 \\
\hline
\hline
\multirow{4}{*}{\centering \rotatebox[origin=c]{90}{\centering Validated \phantom{000000}}} & \raggedright Root & \cellcolor[HTML]{D6C8E8}as of now , snow white lives in the castle . and the princess assures & that the dwarfs \textbf{\color[HTML]{5B3A7A}{\underline{may}}} too . & 40 \\
\cline{2-5}
& \raggedright Epistemic & \cellcolor[HTML]{C5E0D4}in direct sunlight , trolls turn into stone . but the wizard worries & that the dwarfs \textbf{\color[HTML]{3D6B4F}{\underline{may}}} too . & 40 \\
\cline{2-5}
& \raggedright Root & \cellcolor[HTML]{D6C8E8}as of now , snow white lives in the castle . and the princess declares & that the dwarfs \textbf{\color[HTML]{5B3A7A}{\underline{must}}} too . & 40 \\
\cline{2-5}
& \raggedright Epistemic & \cellcolor[HTML]{C5E0D4}in direct sunlight , trolls turn into stone . but the wizard figures & that the dwarfs \textbf{\color[HTML]{3D6B4F}{\underline{must}}} too . & 40 \\
\hline
\hline
\multirow{4}{*}{\centering \rotatebox[origin=c]{90}{\centering Naturalistic}} & \raggedright Root & \cellcolor[HTML]{E8E8E8}{} & you say \textbf{\color[HTML]{5B3A7A}{\underline{may}}} i borrow it please . & 16 \\
\cline{2-5}
& \raggedright Epistemic & \cellcolor[HTML]{E8E8E8}{} & that one \textbf{\color[HTML]{3D6B4F}{\underline{may}}} not be strong enough for you . & 16 \\
\cline{2-5}
& \raggedright Root & \cellcolor[HTML]{E8E8E8}{} & you \textbf{\color[HTML]{5B3A7A}{\underline{must}}} always cut it on the board like that . & 88 \\
\cline{2-5}
& \raggedright Epistemic & \cellcolor[HTML]{E8E8E8}{} & they \textbf{\color[HTML]{3D6B4F}{\underline{must}}} have run away . & 208 \\
\hline
\hline
\multirow{4}{*}{\centering \rotatebox[origin=c]{90}{\centering Naturalistic \scriptsize{(with context)} \phantom{0}}} & \raggedright Root & \cellcolor[HTML]{D6C8E8}you should ask no it is caroline's and you ask & you say \textbf{\color[HTML]{5B3A7A}{\underline{may}}} i borrow it please . & 16 \\
\cline{2-5}
& \raggedright Epistemic & \cellcolor[HTML]{C5E0D4}hm that is not that is not your boat darling your boat is here this is your boat this one & that one \textbf{\color[HTML]{3D6B4F}{\underline{may}}} not be strong enough for you . & 16 \\
\cline{2-5}
& \raggedright Root & \cellcolor[HTML]{D6C8E8}that is it and that one no don't do it on your hands because if it was a real knife you would cut yourself doing that wouldn't you & you \textbf{\color[HTML]{5B3A7A}{\underline{must}}} always cut it on the board like that . & 88 \\
\cline{2-5}
& \raggedright Epistemic & \cellcolor[HTML]{C5E0D4}i can't see them i think they have run away no they are not under there and they are not behind here & they \textbf{\color[HTML]{3D6B4F}{\underline{must}}} have run away . & 208 \\
\hline
\end{tabular}
\caption{Samples of modal expressions with root and epistemic interpretations in different datasets.  EPOS and the validated data are released under a CC BY-SA 4.0 license and the naturalistic (CHILDES) data is released under CC BY-NC-SA 3.0.  Across all datasets, the average sequence length is 15 tokens, with the longest sequence reaching 64 tokens.}
\label{tab:modal_example}
\end{table}

\section{Corpus statistics}

\label{sec:corpus_stats}

\begin{table*}[ht]
\small
\centering
\begin{tabular}{@{}llllrcc@{}}
\toprule
& & & & & \multicolumn{2}{c}{Expectations} \\
\cmidrule(lr){6-7}
Corpus & Modal & Dominant sense & Source & \# & Non-CDL-models & CDL-models\\
\midrule
\multirow{2}{*}{\shortstack{General\\(AmE)}}
& must & root (84\%) & \cite{collins2007} & 930 & root > epistemic & \\
& may & epistemic (83\%) & \cite{hw2012} & 138 & epistemic > root & \\
\midrule
\multirow{2}{*}{\shortstack{CHILDES\\(BrE)\phantom{0000}}}
& must & epistemic (65\%) & \cite{vandooren2022figuring} & 452 & & epistemic > root \\
& may & root (56\%) & \cite{vandooren2022figuring} & 39 & & root > epistemic \\
\bottomrule
\end{tabular}
\caption{Corpus statistics and performance expectations for the dominant senses for modals \textit{may} and \textit{must}.}
\end{table*}

\clearpage

\section{Similarity method plots}
\label{sec:sim_plots}

\begin{figure*}[ht]
    \centering
    \includegraphics[width=0.496\linewidth]{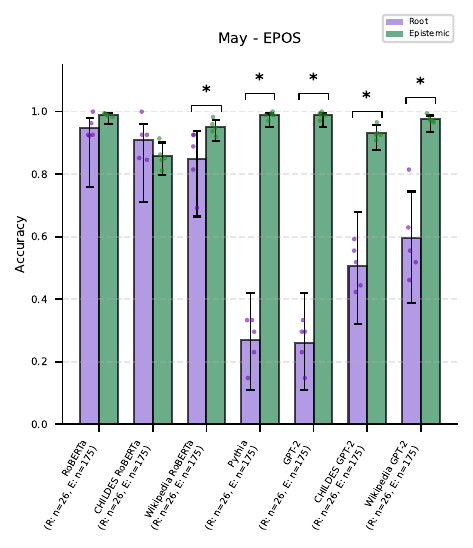}
    \hfill
    \includegraphics[width=0.496\linewidth]{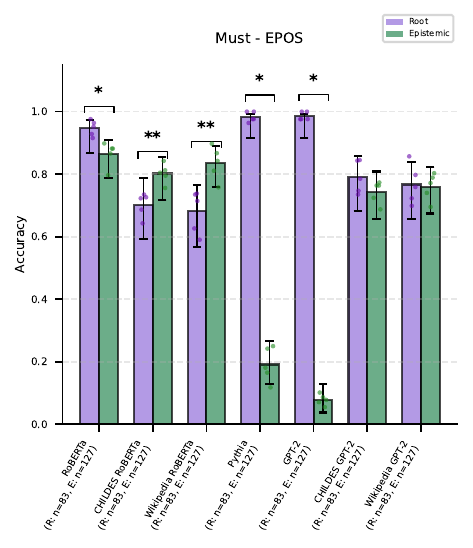}
    \includegraphics[width=0.496\linewidth]{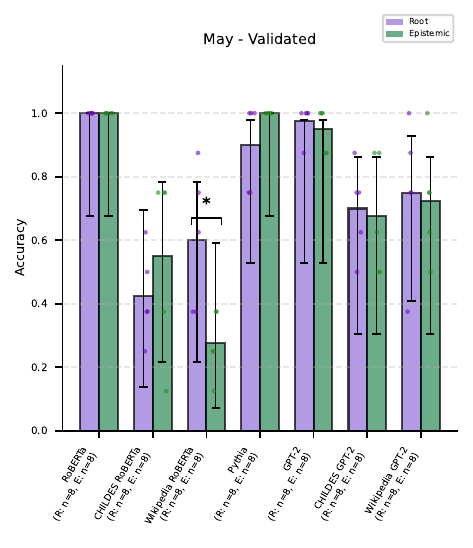}
    \hfill
    \includegraphics[width=0.496\linewidth]{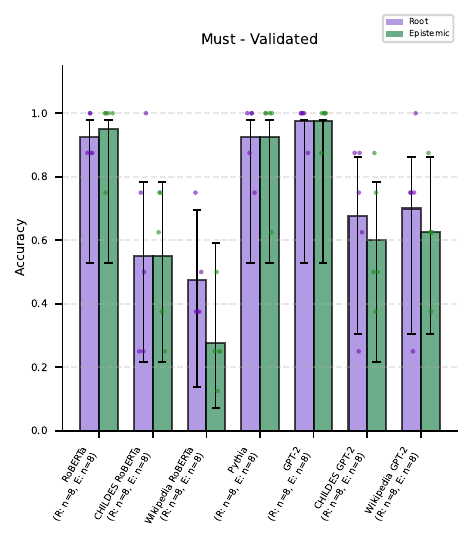}
\end{figure*}

\begin{figure*}[ht]
    \centering
    \includegraphics[width=0.496\linewidth]{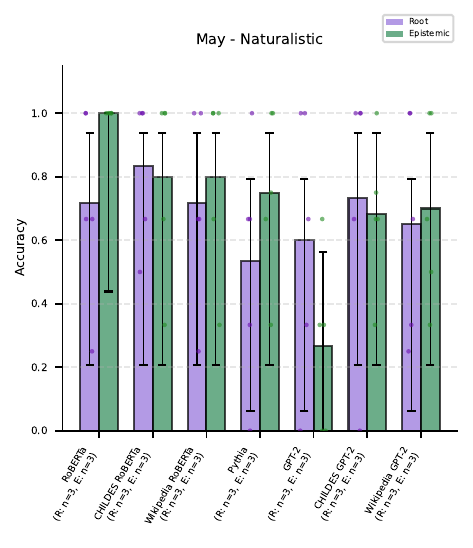}
    \hfill
    \includegraphics[width=0.496\linewidth]{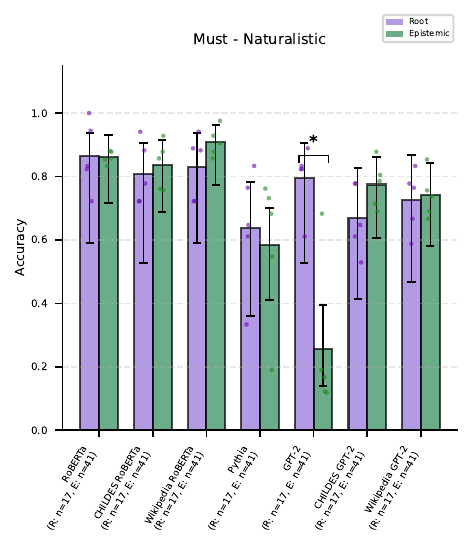}
    \includegraphics[width=0.496\linewidth]{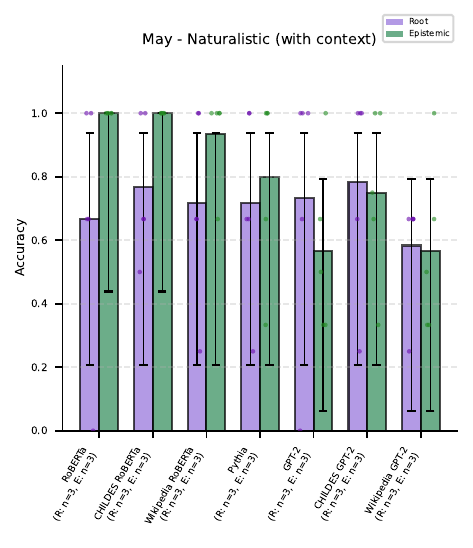}
    \hfill
    \includegraphics[width=0.496\linewidth]{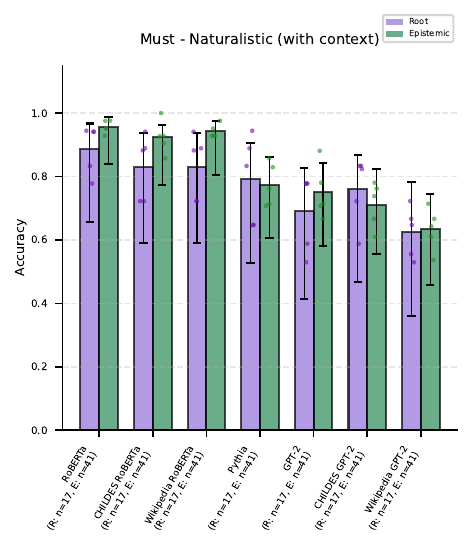}
    \caption{LM accuracy results based on vector similarity scores. Error bars represent binomial 95\% CIs across five cross-validation folds.}
\end{figure*}

\clearpage

\twocolumn

\section{Context variation analysis}
\label{sec:validated_variations}

As a supplementary analysis, we construct two modified evaluation sets based on the original validated dataset from \citet{tulling2020neural} with the goal of investigating the usefulness of different types of linguistic context in modal sense disambiguation. Example data points from the modified datasets can be found in Table \ref{tab:validated_samples}. The first modification removes the preceding sentence that provides the situational background necessary for interpreting the modal. This results in a ``no context'' variant. The second modification targets the clausal structure of the stimuli. In the original dataset, each modal occurs within an embedded clause. The goal behind \citet{tulling2020neural} using this embedded structure is to link the modal interpretation to a third-person entity's perspective, helping bias participants toward a particular modal interpretation in a more clear way. To test whether this perspectival framing itself contributes useful information, we created a ``monoclausal'' variant. In this variant, we remove the matrix clause entirely. This leaves only the embedded clause as an independent sentence. 

We should note that while the original stimuli have been validated with human participants, our modified versions have not. Therefore, we do not know to what extent the intended modal sense distinctions are equally clear in these new variants. Nevertheless, the original dataset was constructed under highly controlled conditions such as balancing for lexical frequency and word lengths across stimuli. These modified datasets, therefore, still provide a controlled setup for isolating the effect of removing different types of context. They also allow us to assess how informative each type of context is for LMs.

\begin{figure*}[ht]
    \centering
    \includegraphics[]{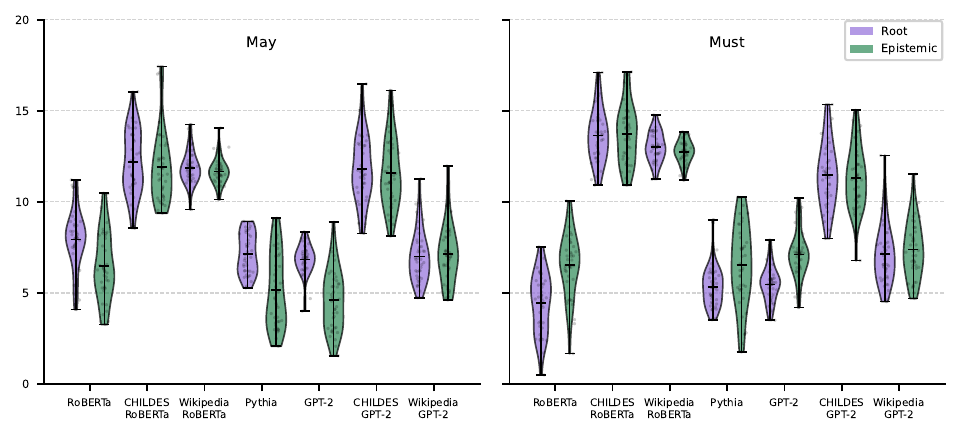}
    \caption{Surprisal score distributions for the modals in validated data. (Figure \ref{fig:validated_surprisal} in the main body.)}
    \label{fig:validated_surprisal_repeat}
\end{figure*}

\begin{figure*}[ht]
    \centering
    \includegraphics[]{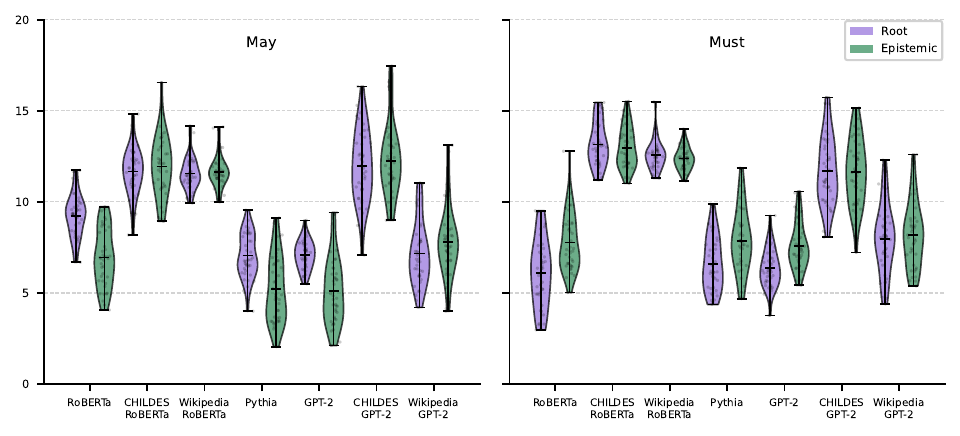}
    \caption{Surprisal score distributions for the modals in validated data (no context).}    \label{fig:validated_surprisal_no_context}
\end{figure*}

\begin{figure*}[ht]
    \centering
    \includegraphics[]{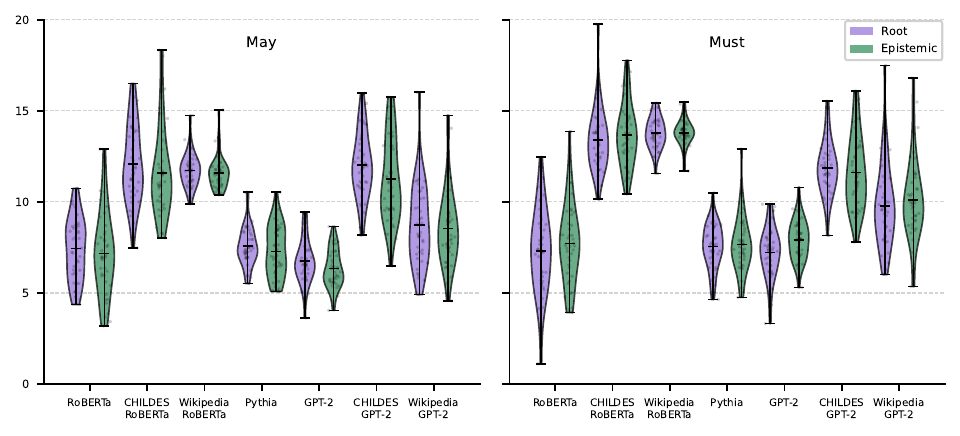}
    \caption{Surprisal score distributions for the modals in validated data (monoclausal).}    \label{fig:validated_surprisal_embedded}
\end{figure*}

We turn first to the surprisal distributions shown in Figures \ref{fig:validated_surprisal_repeat}, \ref{fig:validated_surprisal_no_context}, and \ref{fig:validated_surprisal_embedded}. Removing the situational context alone has a relatively limited impact on how surprisal scores are distributed across modal senses. In both the original and no context conditions, larger LMs maintain a clear surprisal preference for the dominant modal sense. Smaller LMs show little to no disambiguation between root and epistemic interpretations for \textit{may} and \textit{must} in these same conditions.

A different pattern emerges for the monoclausal variant. In this condition, the distribution of surprisal values converges considerably for root and epistemic senses, and this convergence occurs not only for smaller LMs but also for the larger ones. Our earlier experiments show that the larger LMs show a moderate to high capability of distinguishing between modal senses. Given our previous observations, we interpret the convergence in the monoclausal condition as evidence that the removed matrix clause is highly informative for modal sense disambiguation. The removal of the third-person entity prevents LMs from being able to show a clear preference for one modal sense over the other.

We also consider the similarity method results presented in Table \ref{tab:validated_heatmap}. The performance gap between smaller and larger LMs remains wide across the board. For the larger causal LMs, we observe a modest drop in accuracy on the no context variant compared to the original dataset. This drop is followed by a more pronounced drop on the monoclausal variant. The difficulty that LMs exhibit on the monoclausal variant is consistent with their surprisal distributions. We also see that model performance is impacted by the removal of context, which aligns with what we observed in our previous tests using naturalistic data.

Our larger masked LM RoBERTa shows a similar trend for \textit{must}. For \textit{may}, however, the performance on the monoclausal variant is slightly higher than the performance on the no context variant. This difference highlights the benefit of using two different methods to examine model-internal representations, as the consistent pattern we observed regarding modal sense preferences based on surprisal distributions would not have been apparent from the similarity method alone.

Our context-varied results speak to the role of perspective-taking in modal sense disambiguation. In the original stimuli, embedding the target utterance under a matrix clause explicitly attributes the modal judgment to a third-person entity's perspective. When we remove this matrix clause in the monoclausal variant, the utterance is no longer explicitly anchored to a third-person perspective. Both surprisal-based preferences and similarity-based accuracies indicate that the presence of a third-person entity provides a meaningful cue that larger LMs exploit when disambiguating modal senses.

\begin{table*}[ht!]
\centering
\small
\renewcommand{\arraystretch}{2.5}
\setlength{\tabcolsep}{4pt}
\begin{tabular}{|>{\centering\arraybackslash}m{0.7cm}|>{\raggedright\arraybackslash}p{1.3cm}|p{5.8cm}|p{5.75cm}|r|}
\hline
Data & Flavor & Context & Target sequence & \# \\
\hline
\hline
\multirow{4}{*}{\centering \rotatebox[origin=c]{90}{\centering Validated \phantom{000000}}} & \raggedright Root & \cellcolor[HTML]{D6C8E8}as of now , snow white lives in the castle . and the princess assures & that the dwarfs \textbf{\color[HTML]{5B3A7A}{\underline{may}}} too . & 40 \\
\cline{2-5}
& \raggedright Epistemic & \cellcolor[HTML]{C5E0D4}in direct sunlight , trolls turn into stone . but the wizard worries & that the dwarfs \textbf{\color[HTML]{3D6B4F}{\underline{may}}} too . & 40 \\
\cline{2-5}
& \raggedright Root & \cellcolor[HTML]{D6C8E8}as of now , snow white lives in the castle . and the princess declares & that the dwarfs \textbf{\color[HTML]{5B3A7A}{\underline{must}}} too . & 40 \\
\cline{2-5}
& \raggedright Epistemic & \cellcolor[HTML]{C5E0D4}in direct sunlight , trolls turn into stone . but the wizard figures & that the dwarfs \textbf{\color[HTML]{3D6B4F}{\underline{must}}} too . & 40 \\
\hline
\hline
\multirow{4}{*}{\centering \rotatebox[origin=c]{90}{\centering Validated \scriptsize{(no context)} \phantom{0}}} & \raggedright Root & \cellcolor[HTML]{D6C8E8}the princess assures & that the dwarfs \textbf{\color[HTML]{5B3A7A}{\underline{may}}} too . & 40 \\
\cline{2-5}
& \raggedright Epistemic & \cellcolor[HTML]{C5E0D4}the wizard worries & that the dwarfs \textbf{\color[HTML]{3D6B4F}{\underline{may}}} too . & 40 \\
\cline{2-5}
& \raggedright Root & \cellcolor[HTML]{D6C8E8}the princess declares & that the dwarfs \textbf{\color[HTML]{5B3A7A}{\underline{must}}} too . & 40 \\
\cline{2-5}
& \raggedright Epistemic & \cellcolor[HTML]{C5E0D4}the wizard figures & that the dwarfs \textbf{\color[HTML]{3D6B4F}{\underline{must}}} too . & 40 \\
\hline
\hline
\multirow{4}{*}{\centering \rotatebox[origin=c]{90}{\centering Validated \scriptsize{(monoclausal)} \phantom{0}}} & \raggedright Root & \cellcolor[HTML]{D6C8E8}as of now , snow white lives in the castle . & the dwarfs \textbf{\color[HTML]{5B3A7A}{\underline{may}}} too . & 40 \\
\cline{2-5}
& \raggedright Epistemic & \cellcolor[HTML]{C5E0D4}in direct sunlight , trolls turn into stone . & the dwarfs \textbf{\color[HTML]{3D6B4F}{\underline{may}}} too . & 40 \\
\cline{2-5}
& \raggedright Root & \cellcolor[HTML]{D6C8E8}as of now , snow white lives in the castle . & the dwarfs \textbf{\color[HTML]{5B3A7A}{\underline{must}}} too . & 40 \\
\cline{2-5}
& \raggedright Epistemic & \cellcolor[HTML]{C5E0D4}in direct sunlight , trolls turn into stone . & the dwarfs \textbf{\color[HTML]{3D6B4F}{\underline{must}}} too . & 40 \\
\hline
\end{tabular}
\caption{Example data points from different validated dataset versions.}
\label{tab:validated_samples}
\end{table*}

\clearpage

\onecolumn

\begin{table}[ht]
\centering
\small
\renewcommand{\arraystretch}{1.1}
\setlength{\tabcolsep}{4pt}
\begin{tabular}{@{}l c l l *{6}{>{\centering\arraybackslash}p{1.25cm}}@{}}
 & & & & \multicolumn{3}{c}{May} & \multicolumn{3}{c}{Must} \\
\cmidrule(lr){5-7} \cmidrule(lr){8-10}
 & & & & Root & Epistemic & All & Root & Epistemic & All \\
\hline \noalign{\vspace{0.3ex}}
\multirow[l]{9}{*}{\rotatebox{90}{Masked LMs}} & & \multirow[l]{3}{*}{RoBERTa} & Validated & \cellcolor[rgb]{0.39,0.71,0.39} 1.00 & \cellcolor[rgb]{0.39,0.71,0.39} 1.00 & \cellcolor[rgb]{0.39,0.71,0.39} 1.00 & \cellcolor[rgb]{0.48,0.75,0.45} 0.93 & \cellcolor[rgb]{0.45,0.73,0.43} 0.95 & \cellcolor[rgb]{0.47,0.74,0.44} 0.94 \\
 & & & Validated (no context) & \cellcolor[rgb]{0.63,0.82,0.55} 0.80 & \cellcolor[rgb]{0.48,0.75,0.45} 0.93 & \cellcolor[rgb]{0.56,0.78,0.50} 0.86 & \cellcolor[rgb]{0.45,0.73,0.43} 0.95 & \cellcolor[rgb]{0.51,0.76,0.47} 0.90 & \cellcolor[rgb]{0.48,0.75,0.45} 0.93 \\
 & & & Validated (monoclausal) & \cellcolor[rgb]{0.57,0.79,0.51} 0.85 & \cellcolor[rgb]{0.48,0.75,0.45} 0.93 & \cellcolor[rgb]{0.53,0.77,0.48} 0.89 & \cellcolor[rgb]{0.54,0.78,0.49} 0.88 & \cellcolor[rgb]{0.57,0.79,0.51} 0.85 & \cellcolor[rgb]{0.56,0.78,0.50} 0.86 \\
\cline{3-10} \noalign{\vspace{0.4ex}}
 & & \multirow[l]{3}{*}{\shortstack[l]{CHILDES\\RoBERTa}} & Validated & \cellcolor[rgb]{1,0.85,0.67} 0.42 & \cellcolor[rgb]{0.94,0.97,0.74} 0.55 & \cellcolor[rgb]{1,0.97,0.76} 0.49 & \cellcolor[rgb]{0.94,0.97,0.74} 0.55 & \cellcolor[rgb]{0.94,0.97,0.74} 0.55 & \cellcolor[rgb]{0.94,0.97,0.74} 0.55 \\
 & & & Validated (no context) & \cellcolor[rgb]{1,1.00,0.78} 0.50 & \cellcolor[rgb]{0.94,0.97,0.74} 0.55 & \cellcolor[rgb]{0.97,0.98,0.76} 0.53 & \cellcolor[rgb]{0.85,0.93,0.69} 0.62 & \cellcolor[rgb]{1,0.80,0.63} 0.40 & \cellcolor[rgb]{0.98,0.99,0.77} 0.51 \\
 & & & Validated (monoclausal) & \cellcolor[rgb]{1,0.80,0.63} 0.40 & \cellcolor[rgb]{0.97,0.98,0.76} 0.53 & \cellcolor[rgb]{1,0.92,0.73} 0.46 & \cellcolor[rgb]{1,0.90,0.71} 0.45 & \cellcolor[rgb]{0.94,0.97,0.74} 0.55 & \cellcolor[rgb]{1,1.00,0.78} 0.50 \\
\cline{3-10} \noalign{\vspace{0.4ex}}
 & & \multirow[l]{3}{*}{\shortstack[l]{Wikipedia\\RoBERTa}} & Validated & \cellcolor[rgb]{0.88,0.94,0.71} 0.60 & \cellcolor[rgb]{1,0.55,0.43} 0.28 & \cellcolor[rgb]{1,0.87,0.69} 0.44 & \cellcolor[rgb]{1,0.95,0.75} 0.47 & \cellcolor[rgb]{1,0.55,0.43} 0.28 & \cellcolor[rgb]{1,0.75,0.59} 0.38 \\
 & & & Validated (no context) & \cellcolor[rgb]{0.78,0.89,0.64} 0.68 & \cellcolor[rgb]{0.76,0.88,0.63} 0.70 & \cellcolor[rgb]{0.77,0.89,0.64} 0.69 & \cellcolor[rgb]{0.73,0.87,0.61} 0.72 & \cellcolor[rgb]{0.82,0.91,0.67} 0.65 & \cellcolor[rgb]{0.77,0.89,0.64} 0.69 \\
 & & & Validated (monoclausal) & \cellcolor[rgb]{0.88,0.94,0.71} 0.60 & \cellcolor[rgb]{1,0.85,0.67} 0.42 & \cellcolor[rgb]{0.98,0.99,0.77} 0.51 & \cellcolor[rgb]{0.88,0.94,0.71} 0.60 & \cellcolor[rgb]{1,0.85,0.67} 0.42 & \cellcolor[rgb]{0.98,0.99,0.77} 0.51 \\
\hline \noalign{\vspace{0.3ex}}
\multirow[l]{12}{*}{\rotatebox{90}{Causal LMs}} & & \multirow[l]{3}{*}{Pythia} & Validated & \cellcolor[rgb]{0.51,0.76,0.47} 0.90 & \cellcolor[rgb]{0.39,0.71,0.39} 1.00 & \cellcolor[rgb]{0.45,0.73,0.43} 0.95 & \cellcolor[rgb]{0.48,0.75,0.45} 0.93 & \cellcolor[rgb]{0.48,0.75,0.45} 0.93 & \cellcolor[rgb]{0.48,0.75,0.45} 0.93 \\
 & & & Validated (no context) & \cellcolor[rgb]{0.54,0.78,0.49} 0.88 & \cellcolor[rgb]{0.45,0.73,0.43} 0.95 & \cellcolor[rgb]{0.50,0.76,0.46} 0.91 & \cellcolor[rgb]{0.48,0.75,0.45} 0.93 & \cellcolor[rgb]{0.57,0.79,0.51} 0.85 & \cellcolor[rgb]{0.53,0.77,0.48} 0.89 \\
 & & & Validated (monoclausal) & \cellcolor[rgb]{0.66,0.84,0.57} 0.78 & \cellcolor[rgb]{0.60,0.81,0.53} 0.82 & \cellcolor[rgb]{0.63,0.82,0.55} 0.80 & \cellcolor[rgb]{0.63,0.82,0.55} 0.80 & \cellcolor[rgb]{0.63,0.82,0.55} 0.80 & \cellcolor[rgb]{0.63,0.82,0.55} 0.80 \\
\cline{3-10} \noalign{\vspace{0.4ex}}
 & & \multirow[l]{3}{*}{GPT-2} & Validated & \cellcolor[rgb]{0.42,0.72,0.41} 0.97 & \cellcolor[rgb]{0.45,0.73,0.43} 0.95 & \cellcolor[rgb]{0.44,0.73,0.42} 0.96 & \cellcolor[rgb]{0.42,0.72,0.41} 0.97 & \cellcolor[rgb]{0.42,0.72,0.41} 0.97 & \cellcolor[rgb]{0.42,0.72,0.41} 0.97 \\
 & & & Validated (no context) & \cellcolor[rgb]{0.45,0.73,0.43} 0.95 & \cellcolor[rgb]{0.54,0.78,0.49} 0.88 & \cellcolor[rgb]{0.50,0.76,0.46} 0.91 & \cellcolor[rgb]{0.48,0.75,0.45} 0.93 & \cellcolor[rgb]{0.54,0.78,0.49} 0.88 & \cellcolor[rgb]{0.51,0.76,0.47} 0.90 \\
 & & & Validated (monoclausal) & \cellcolor[rgb]{0.57,0.79,0.51} 0.85 & \cellcolor[rgb]{0.57,0.79,0.51} 0.85 & \cellcolor[rgb]{0.57,0.79,0.51} 0.85 & \cellcolor[rgb]{0.63,0.82,0.55} 0.80 & \cellcolor[rgb]{0.69,0.85,0.59} 0.75 & \cellcolor[rgb]{0.66,0.84,0.57} 0.78 \\
\cline{3-10} \noalign{\vspace{0.4ex}}
 & & \multirow[l]{3}{*}{\shortstack[l]{CHILDES\\GPT-2}} & Validated & \cellcolor[rgb]{0.76,0.88,0.63} 0.70 & \cellcolor[rgb]{0.78,0.89,0.64} 0.68 & \cellcolor[rgb]{0.77,0.89,0.64} 0.69 & \cellcolor[rgb]{0.78,0.89,0.64} 0.68 & \cellcolor[rgb]{0.88,0.94,0.71} 0.60 & \cellcolor[rgb]{0.83,0.92,0.67} 0.64 \\
 & & & Validated (no context) & \cellcolor[rgb]{0.82,0.91,0.67} 0.65 & \cellcolor[rgb]{0.76,0.88,0.63} 0.70 & \cellcolor[rgb]{0.78,0.89,0.64} 0.68 & \cellcolor[rgb]{1,1.00,0.78} 0.50 & \cellcolor[rgb]{0.85,0.93,0.69} 0.62 & \cellcolor[rgb]{0.92,0.96,0.73} 0.56 \\
 & & & Validated (monoclausal) & \cellcolor[rgb]{0.82,0.91,0.67} 0.65 & \cellcolor[rgb]{0.88,0.94,0.71} 0.60 & \cellcolor[rgb]{0.85,0.93,0.69} 0.62 & \cellcolor[rgb]{0.78,0.89,0.64} 0.68 & \cellcolor[rgb]{0.94,0.97,0.74} 0.55 & \cellcolor[rgb]{0.86,0.93,0.69} 0.61 \\
\cline{3-10} \noalign{\vspace{0.4ex}}
 & & \multirow[l]{3}{*}{\shortstack[l]{Wikipedia\\GPT-2}} & Validated & \cellcolor[rgb]{0.69,0.85,0.59} 0.75 & \cellcolor[rgb]{0.73,0.87,0.61} 0.72 & \cellcolor[rgb]{0.71,0.86,0.60} 0.74 & \cellcolor[rgb]{0.76,0.88,0.63} 0.70 & \cellcolor[rgb]{0.85,0.93,0.69} 0.62 & \cellcolor[rgb]{0.80,0.90,0.65} 0.66 \\
 & & & Validated (no context) & \cellcolor[rgb]{0.63,0.82,0.55} 0.80 & \cellcolor[rgb]{0.73,0.87,0.61} 0.72 & \cellcolor[rgb]{0.68,0.84,0.58} 0.76 & \cellcolor[rgb]{0.73,0.87,0.61} 0.72 & \cellcolor[rgb]{0.88,0.94,0.71} 0.60 & \cellcolor[rgb]{0.80,0.90,0.65} 0.66 \\
 & & & Validated (monoclausal) & \cellcolor[rgb]{0.94,0.97,0.74} 0.55 & \cellcolor[rgb]{0.66,0.84,0.57} 0.78 & \cellcolor[rgb]{0.80,0.90,0.65} 0.66 & \cellcolor[rgb]{0.94,0.97,0.74} 0.55 & \cellcolor[rgb]{0.78,0.89,0.64} 0.68 & \cellcolor[rgb]{0.86,0.93,0.69} 0.61 \\
\hline
\end{tabular}
\caption{LM accuracy results on variations of validated data.}
\label{tab:validated_heatmap}
\end{table}

\vspace{1.5cm}

\end{document}